\documentclass[fleqn,10pt]{wlscirep}
\usepackage[utf8]{inputenc}
\usepackage[T1]{fontenc}
\title{Maia: A Real-time Nonverbal Chat for Human-AI Interaction}

\author[1, +]{Dragos Costea}
\author[2, +]{Alina Marcu}
\author[3]{Cristina Lazar}
\author[1,2,*]{Marius Leordeanu}
\affil[1]{University Politehnica of Bucharest, 313 Splaiul Independentei, Bucharest, Romania}
\affil[2]{Institute of Mathematics of the Romanian Academy, 21 Calea Grivitei, Bucharest, Romania}
\affil[3]{National University of Arts, 19 Constantin Budisteanu, Bucharest, Romania}

\affil[*]{Primary contact: Marius Leordeanu -- leordeanu@gmail.com}

\affil[+]{these authors contributed equally to this work}

\begin{abstract}

Modeling face-to-face communication in computer vision, which focuses on recognizing and analyzing nonverbal cues and behaviors during interactions, serves as the foundation for our proposed alternative to text-based Human-AI interaction. By leveraging nonverbal visual communication, through facial expressions, head and body movements, we aim to enhance engagement and capture the user's attention through a novel improvisational element, that goes beyond mirroring gestures. Our goal is to track and analyze facial expressions, and other nonverbal cues in real-time, and use this information to build models that can predict and understand human behavior. Operating in real-time and requiring minimal computational resources, our approach signifies a major leap forward in making AI interactions more natural and accessible. We offer three different complementary approaches, based on retrieval, statistical, and deep learning techniques. A key novelty of our work is the integration of an artistic component atop an efficient human-computer interaction system, using art as a medium to transmit emotions. Our approach is not art-specific and can be adapted to various paintings, animations, and avatars. In our experiments, we compare state-of-the-art diffusion models as mediums for emotion translation in 2D, and our 3D avatar, Maia, that we introduce in this work, with not just facial movements but also body motions for a more natural and engaging experience. We demonstrate the effectiveness of our approach in translating AI-generated emotions into human-relatable expressions, through both human and automatic evaluation procedures, highlighting its potential to significantly enhance the naturalness and engagement of Human-AI interactions across various applications.

\end{abstract}

\begin{document}

\flushbottom
\maketitle
%
%
\thispagestyle{empty}

\section*{Introduction}
\vspace{2mm}
In the evolving landscape of Human-Machine Interfacing (HMI), the quest for more intuitive and human-like interactions has led to groundbreaking research in face-to-face communication modeling. Accurate detection of nonverbal facial expressions that represent emotions~\cite{cowen2020face, davidson2012emotional, barrett2016works}
constitutes a critical element in the advancement of technologies capable of engaging in seamless conversational interactions with humans. nonverbal cues can transmit more information regarding emotions, through expressions, behavior, and appearance, in a universal language especially using the wide range of multimedia data available nowadays~\cite{zhao2019affective}. The task of detecting and most importantly responding to such nonverbal expressions, in order to engage in interaction, has been largely overlooked by researchers, likely due to the limited availability of data capturing a sufficiently wide variety of such peer-to-peer interactions. Recent works focus on using audio cues or vocalizations~\cite{tzirakis2023large} to understand the user's emotion, but in terms of response generation, most methods model facial movements in the context of talking-head generation in the image space~\cite{gupta2023towards, zhang2023sadtalker, agarwal2023audio} or animated drawings~\cite{smith2023method} using facial keypoints. 

\vspace{1mm}
\noindent Human-AI interaction has seen its most successful adoption and promising results in terms of chat~\cite{bubeck2023sparks}, but nonverbal chat systems using facial emotions have received little to no attention. There are systems for generating gestures from speech~\cite{li2021audio2gestures, ghorbani2023zeroeggs, luo2023reactface}, challenges and datasets for reactive heads ~\cite{song2024react, ren2024veatic} or even full body animation as interaction response~\cite{raimbaud2021reactive}, but most are limited in scope (e.g, virtual interviews~\cite{anderson2013tardis, busso2008iemocap} or conversation~\cite{song2023emotional}) and do not feature a genuine response based on nonverbal interaction only. The Smile Project Deep Immersive Art with Real-time Human-AI Interaction, introduced in \cite{smileproject}, is one of the first such systems that provide real-time feedback with light and color to the user's smile and also gestures and pose. 

\vspace{1mm}
\noindent We encountered several significant challenges while developing our approach for efficient Human-AI nonverbal interaction:
\vspace{-1mm}
\begin{itemize}
    \item \textbf{Complexity of nonverbal communication:} Accurately processing fine-grained facial expressions without any verbal cues is inherently difficult, as nonverbal communication involves subtle and intricate signals.
    \vspace{-1mm}
    \item \textbf{Individual variability:} Each person possesses a unique repertoire of expressions for different emotions, making it difficult to develop a universal algorithm that produces convincing results across diverse populations.
    \vspace{-1mm}
    \item \textbf{Real-time processing demands:} For natural conversation, interaction must be continuous. Even a brief delay of a few seconds in emotion recognition and response can lead to misinterpretation, potentially undermining the efficacy of the communication channel.
    \vspace{-1mm}
    \item \textbf{Limited prior research:} The scarcity of relevant data and established procedures posed significant obstacles. While extensive research exists on modeling talking heads using audio or verbal cues, resulting in numerous conversational datasets \cite{zhou2022responsive, vicox, poria2018meld, zafeiriou2017aff}, there is a notable lack of datasets and methodologies focused exclusively on behavioral gestures in nonverbal human-AI interaction.
\end{itemize}

\vspace{1mm}
\noindent In this paper, we explore multiple methods in the image space, in the facial and body keypoints space (both at input and output) and even 3D with multiple mediums to rely on emotion.
In the keypoints space, we use a facial keypoints temporal sequence to generate a new temporal keypoints sequence. While there are methods that generate and match keypoints from a set of images (e.g.,~\cite{jiang2021cotr}), to the best of our knowledge, our work is the first to propose such a pipeline in the context of nonverbal chat, seamless, without the need of labor-intensive human annotations, even though learning is involved and we are also among the first to propose a novel dataset for this task.

\section*{Prior work}
\vspace{2mm}
\noindent\textbf{Methods} --  In 2008, the work of~\cite{cristescu2008emotions} explored the role of nonverbal behaviors, such as facial expressions and body movements, in conveying emotions during human-computer interactions (HCI). Using psychological theories and computational models, the study demonstrated that nonverbal cues significantly enhance the emotional dynamics of HCI, making interactions more responsive and empathetic. This early work laid the foundation for understanding the impact of nonverbal communication on user experience. Following this,~\cite{lin2011nonverbal} focused on nonverbal acoustic signals like intonation and rhythm, utilizing signal processing and machine learning to analyze their impact on communication efficacy. The study found that these acoustic signals significantly enhance the clarity and effectiveness of communication. This research highlighted the importance of acoustic signals, providing a new dimension to nonverbal communication in HCI. 

\vspace{1mm}
\noindent In 2016,~\cite{jofre2016non} investigated integrating nonverbal cues into virtual reality (VR) interfaces using motion capture and gesture recognition. Their findings showed that these cues significantly enhance user engagement and interaction quality in VR environments. This study demonstrated the practical applications of nonverbal communication in VR, making interactions more intuitive and immersive. In 2019,~\cite{gon2019machine} applied machine learning techniques to interpret nonverbal cues in HCI. Utilizing Convolutional Neural Networks (CNNs) and Recurrent Neural Networks (RNNs), the study analyzed facial expressions, gestures, and body language, showing high accuracy in recognizing and interpreting these cues. This research showcased the effectiveness of advanced machine learning models in enhancing the interpretation of nonverbal communication. In the same year,~\cite{saunderson2019robots} reviewed existing research on the impact of nonverbal communication in social human-robot interaction (HRI), categorizing various nonverbal behaviors and their effects on human perceptions and interactions with robots. This comprehensive survey provided a valuable overview of nonverbal communication in HRI, highlighting its significance in improving robot social presence and acceptance.

\vspace{1mm}
\noindent Further advancing the field, the work from~\cite{raimbaud2021reactive} proposed a viewpoint-driven approach for virtual agents to use nonverbal communication based on the user's perspective. The study found that adapting nonverbal behaviors dynamically improves the effectiveness of virtual agents in communication. This novel approach emphasized the importance of considering user perspective to enhance context-awareness in nonverbal communication. In 2023,~\cite{urakami2023nonverbal} examined the integration of nonverbal communication cues into HRI from a communication studies perspective, offering a framework based on literature from communication studies. The study proposed design patterns for robotic nonverbal behavior to enhance social presence. This work bridged communication studies and HRI, providing a comprehensive framework for integrating nonverbal cues. Further work determined that the impact on non-verbal first impression influenced significantly the interaction time with the agent~\cite{cafaro2016first}.

\vspace{1mm}
\noindent Recently,~\cite{lozano2023open} introduced an open framework for integrating nonverbal communication into HRI, validated using multiple datasets. The framework proved effective in enhancing human-robot interactions through improved understanding and generation of nonverbal behaviors. The evolution of nonverbal communication in HCI and HRI highlights its critical role in enhancing interactive systems.

\vspace{1mm}
\noindent\textbf{Datasets} -- To the best of our knowledge, there are no public nonverbal facial response datasets available. Most of the available ones, such as MELD~\cite{poria2018meld}, consider the emotion as a result of a verbal cue, with facial expressions serving mostly as an enhancement of the speech, but not as a main cause. This work investigates the latter, where an emotion results from another emotion, without any speech contribution. While there are a number of works proposing dyadic nonverbal interaction models, such as IEMOCAP~\cite{busso2008iemocap}, they do not develop a practical response system. In 2023,~\cite{johnson2023recanvo} introduced the ReCANVo dataset, a database of real-world communicative and affective nonverbal vocalizations. Although not directly related to our work, this dataset exemplifies existing datasets using speech/vocalizations, emphasizing the prevalent focus on speech and underscoring the need to use motion as a form of emotional expression, not just vocalizations, particularly in studies linked to autism. Additionally, the nonverbal Vocalization Dataset~\cite{deeplyinc_nonverbal_2021} and the Parent-Child-Vocal-Interaction-Dataset~\cite{deeplyinc_parentchild_2021} provide further examples of datasets emphasizing vocal interactions.

\subsection*{Main contributions} 
\vspace{1mm}
We make pivotal contributions to the field of Human-AI communication, offering: 
\begin{itemize} 
    \item The first-of-its-kind unsupervised learning solution that leverages visual and artistic cues for Human-AI communication, capable of expressing emotions in real-time with minimal computational resources. This approach not only lowers the barrier for real-time emotional communication between humans and AI but also translates complex emotions into a more universally understandable language.
    \item Compelling results in emotion expressivity through multiple mediums (here medium as a means of communication), based on the evaluation of both humans as well as automatic emotion recognition procedures.
    \item A comprehensive dataset made in collaboration with the National University of Arts in Bucharest, first of its kind for nonverbal communication that encompasses a wide range of nonverbal expressions, meticulously curated to facilitate the development of advanced models for emotion recognition and interaction.
\end{itemize}

\noindent Through these contributions, we aim to catalyze further innovation in the field, encouraging researchers to explore new dimensions of non-verbal communication and develop solutions that are increasingly empathetic and adept at conveying emotional nuances.

\section*{Nonverbal Facial Response}
\vspace{2mm}
The current section outlines our methodological advances in enabling AI systems to recognize and generate nonverbal facial responses in real-time. We leveraged advanced computer vision techniques to accurately interpret and mimic human emotions through facial expressions and body movement, enhancing the AI's capability to engage in a manner that mimics natural human interaction in the 2D (painting) and 3D (animated avatar) spectrum, but using the temporal component for both in video for a more natural response. 



\subsection*{Nonverbal Expressions Dataset (NED)}
\vspace{2mm}
Due to the scarcity of nonverbal expression transmission datasets, we have collected our own, which we will make publicly available for further research. For each type of emotion from a predefined set, we collect 30 videos with expressions depicting the emotion, in a controlled environment --- from frame-centered users, facing the camera, ranging from 3 up to 6 seconds per video session. The users have expressed consent to use their data in our experiments for research purposes. All videos were collected at a frame rate of 30 FPS. Inspired by available datasets for emotion recognition, we collected videos depicting mostly positive emotions~\textit{charming, happy, impressed, neutral, shocked, surprised, laughing}. In our experimental analysis, we prioritized the positive emotions due to our end goal of expressing positivity, therefore for our current experiments, we sampled the videos depicting~\textit{happy, laughing and surprised} expressions, exclusively for the expression generation experiments. 
For all our experiments we split the proposed NED dataset for training and testing with a 80 - 20\% ratio for each emotion. Each one of our methods outputs the same number of frames as it receives as input. 

\subsection*{Introducing Maia}
\begin{figure}[ht!]
  \centering
  \includegraphics[width=\linewidth]{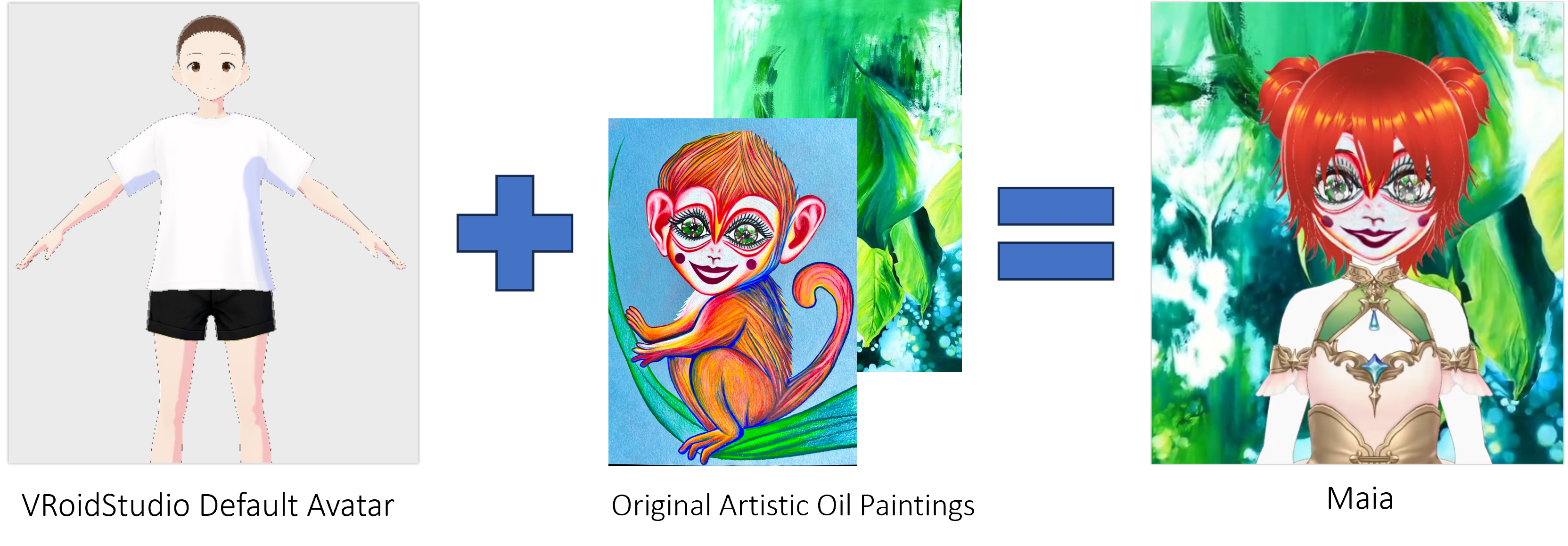}
  \caption{The process of creating our 3D avatar Maia. We used a stadard VRoidStudio feminine avatar on top of which we applied the facial texture and the eyes from the painting on top of the avatar's face matching each correspondence manually and then customizing her with hair and clothing based on her personality. We used a different painting as the background for Maia.}
\label{fig:Maia}
\vspace{-1em}
\end{figure}

\begin{figure*}[ht!]
\centering
\includegraphics[width=\linewidth]{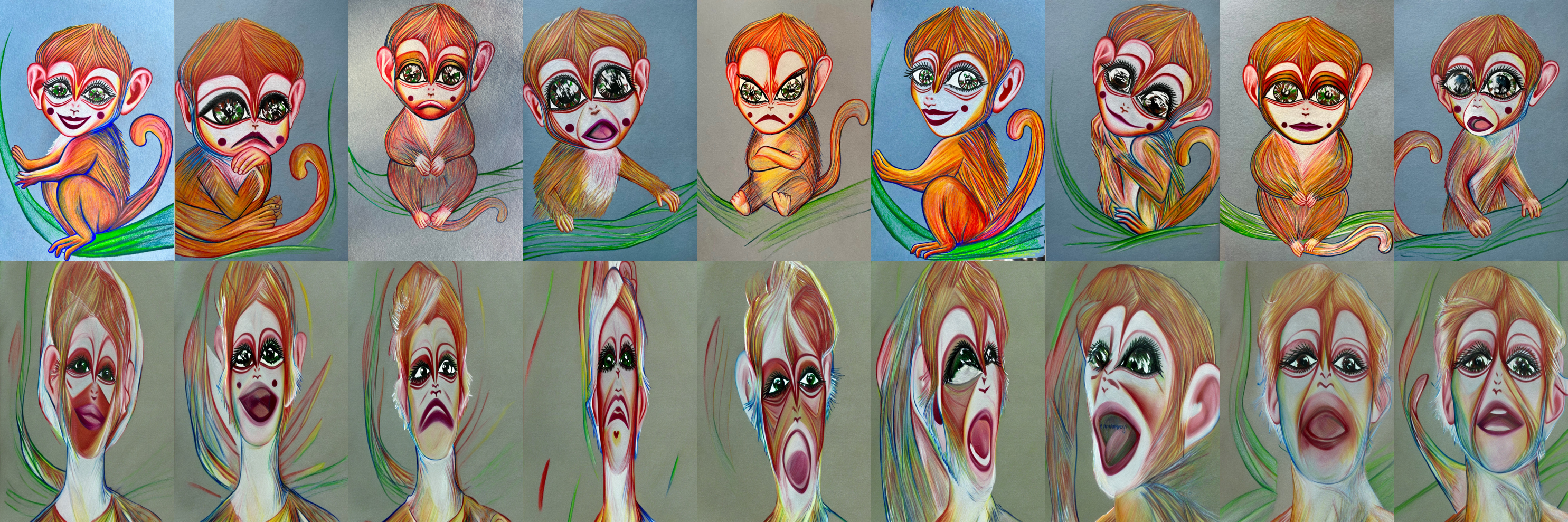}
\caption{Evolution of Maia using 2D generation methods with diffusion. The first row depicts original paintings portraying a wide range of emotions such as happy, sad, surprised, shocked or angry. The second row represents some qualitative results using a video with a human portraying different emotions on top of which we apply the style of the paintings (the diffusion model fine-tuned on the original paintings using LoRA).}
\label{fig:maia_evolution}
\vspace{-1em}
\end{figure*}

\vspace{2mm}
Our work introduces several keypoints-to-keypoints transformation methods, which are easy to integrate with many animation software solutions. However, we are concerned with the way humans interact with a character. Therefore, we created our character to animate based on our keypoints (see Figure~\ref{fig:Maia}). We introduce Maia, an animated character that was converted to 3D using VRoidStudio~\cite{isozaki2021vroid} and several original artistic oil paintings made by one of our artists. For animation, we use VSeeFace~\cite{VSeeFace}. The texture used for creating our avatar is the first painting in the first row from Figure~\ref{fig:maia_evolution}, depicting the emotion \textit{happy}. Employing such a custom 3D avatar with an applied artistic texture derived from a painting offers a compelling and innovative approach to avatar design. The integration of beloved art pieces enhances user immersion, positively impacting mood and emotional connection. Furthermore, this combination of art and technology can enhance user experiences, foster artistic collaboration, and open new opportunities for creative expression in the digital realm.

\subsection*{Evaluation Procedures}
\vspace{2mm}
To properly assess the effectiveness of our methods in conveying natural emotions and expressivity, we employ two means of evaluation: an automatic evaluation using GPT-4 and a human-level evaluation.

\vspace{2mm}
\noindent\textbf{Automatic Evaluation} -- Our evaluation method leverages GPT-4o~\cite{openai2024gpt4}, one of the most advanced pre-trained models available, to classify emotions in individual video frames from our test set sequences. The process begins with the extraction of individual frames from these test videos. GPT-4 then analyzes each frame, assigning emotion labels based on detected facial expressions and contextual cues. This crucial step transforms raw visual data into meaningful emotion predictions, employing advanced machine learning techniques to decode complex emotional signals. Subsequently, we aggregate the classified emotions from all frames within a sequence to determine the dominant emotion (majority). This aggregation captures the overall emotional tone, ensuring that transient expressions or brief anomalies do not skew the results. By summarizing emotions across multiple frames, we obtain a holistic view of the emotional content, reflecting the general sentiment throughout the sequence. For this task, we provided GPT-4 with a specific system prompt: \textit{"You are a helpful assistant generating a video emotion summary from 10 frames of the same video. Classify each frame into one of three emotions: happy, surprised, or laughing. Provide an output for each frame."} We have modified the prompt for each of our experiments (3 or 5 emotions) accordingly. This approach allows for a comprehensive yet focused analysis of emotional content in video sequences, harnessing the power of advanced AI to interpret complex visual data. 

\vspace{2mm}
\noindent\textbf{Human-level Evaluation} -- To complement our automatic evaluation, we conducted a comprehensive human-level assessment to capture nuanced aspects of emotional expression that may elude AI-based systems. We recruited a group of individuals to evaluate the same test set as the one used in the automatic evaluation for fair comparison. Each annotator provided a class per video after which we evaluated the performance based on accuracy. In our tables, for simplicity, we opted to average all the performances and report them as Humans (Mean). 


\subsection*{2D generation with diffusion}
\vspace{2mm}
\noindent Stable diffusion~\cite{rombach2021highresolution} emerged as a text-to-image synthetic image generator that later evolved with the help of fast fine-tuning tools such as Dreambooth~\cite{ruiz2023dreambooth} and preconditioning tools such as ControlNet~\cite{zhang2023adding} into a highly customizable and high-quality generation pipeline. Our 2D approach leverages advanced diffusion models and computer vision techniques to translate human emotions into artistic representations. The core of our approach begins with input processing, where we use OpenCV to capture and manipulate video input on a frame-by-frame basis, focusing on relevant facial regions through strategic cropping. To extract crucial data about body positioning and movement, we employ the OpenPose detector~\cite{cao2019openpose}. This pose estimation step provides a foundation for understanding the subject's nonverbal cues. Complementing this, we utilize a depth estimation pipeline~\cite{young2023depth} to generate depth maps for each frame, adding a dimension of spatial information to our input.

\begin{table}[!ht]
    \setlength{\tabcolsep}{2.5pt} 
    \centering
    \begin{tabular}{lcc}
        \toprule
        Set & Humans (Mean) & GPT-4o\\
        \toprule
        3 emotions & \textbf{91.66} & 55.56 \\
        5 emotions & \textbf{50.69} & 47.22 \\
        \bottomrule
    \end{tabular} 
    \caption{Human and machine-level evaluation for 3 and 5 emotions generated with DreamBooth and ControlNet. We present the accuracy over all classes in percentages for each of our human evaluators and after computing the majority score based on our experts. The automatic methods produce similar results compared to human annotators in the more difficult case of 5 emotions. For a detailed analysis of these results, we present the confusion matrices in Figure~\ref{fig:confusion_matrices_2D_3_and_5emotions}.}
    \label{tab:human-eval-diffusion}
\vspace{-1em}
\end{table}

\vspace{2mm}
\noindent At the heart of our image generation process lies a Stable Diffusion XL model~\cite{rombach2022high} with dual ControlNet~\cite{zhang2023controlnet} conditioning. The ControlNet modules, specifically trained for depth and pose control, allow for fine-grained manipulation of the generated output. To ensure consistent stylistic output across generated frames, we incorporate a custom LoRA~\cite{hu2022lora} weight file, fine-tuned on a specific artistic style or character (in this case, a "TOK monkey"). The stable diffusion model was fine-tuned using all the paintings of varied emotions with Maia (see Figure~\ref{fig:maia_evolution} first row). The diffusion process is guided by carefully crafted prompts, with negative prompts helping to avoid common artifacts and quality issues. Our system processes videos in batches, organized by emotion categories, allowing for systematic analysis across different emotional states. For each input frame, we generate a corresponding artistic interpretation, saving it as a high-quality PNG image.


\begin{figure}[!ht]
    \begin{minipage}[t]{0.52\linewidth}
        \includegraphics[width=\linewidth]{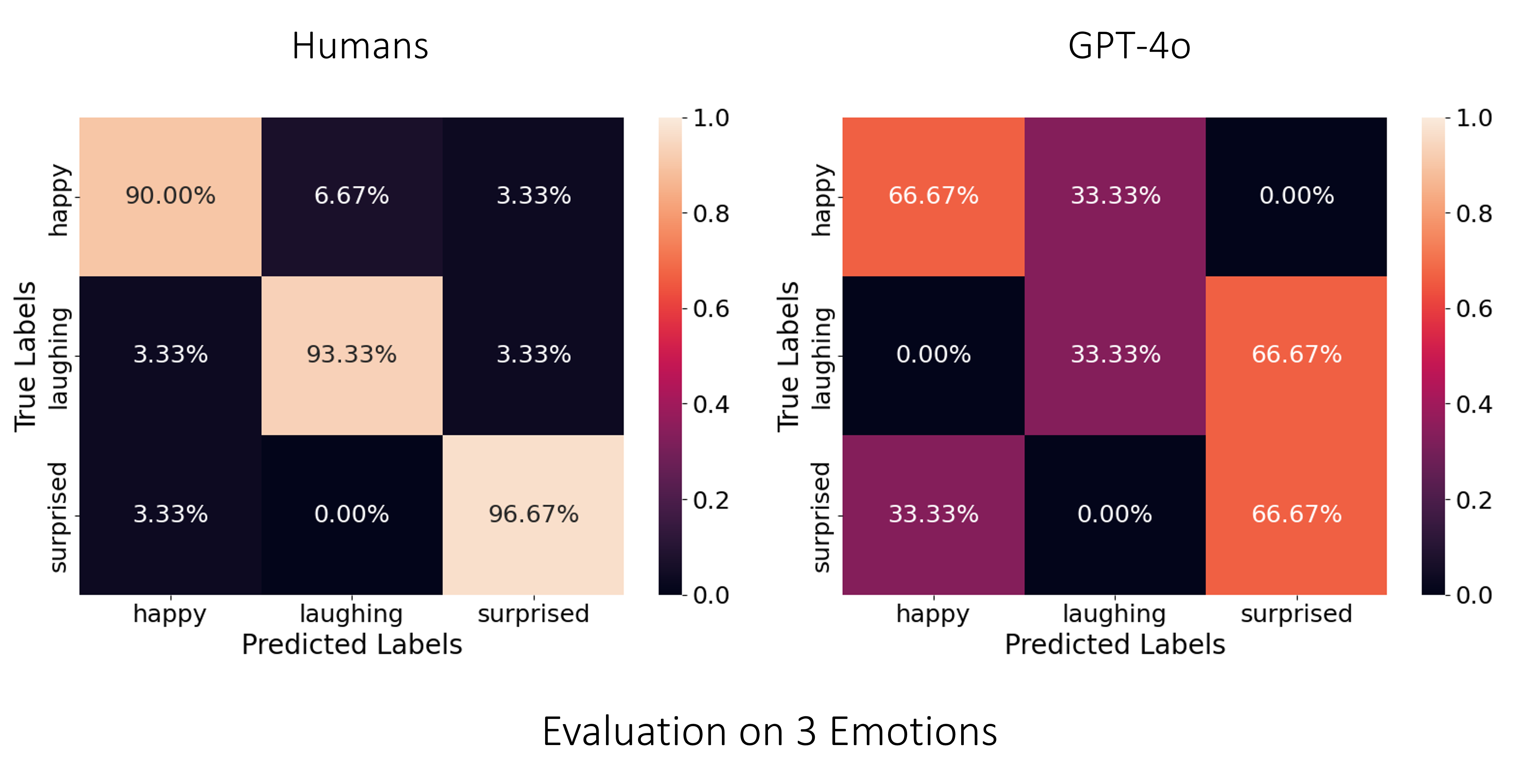}
    \end{minipage}
    \hspace{-1em}
    \begin{minipage}[t]{0.52\linewidth}
        \includegraphics[width=\linewidth]{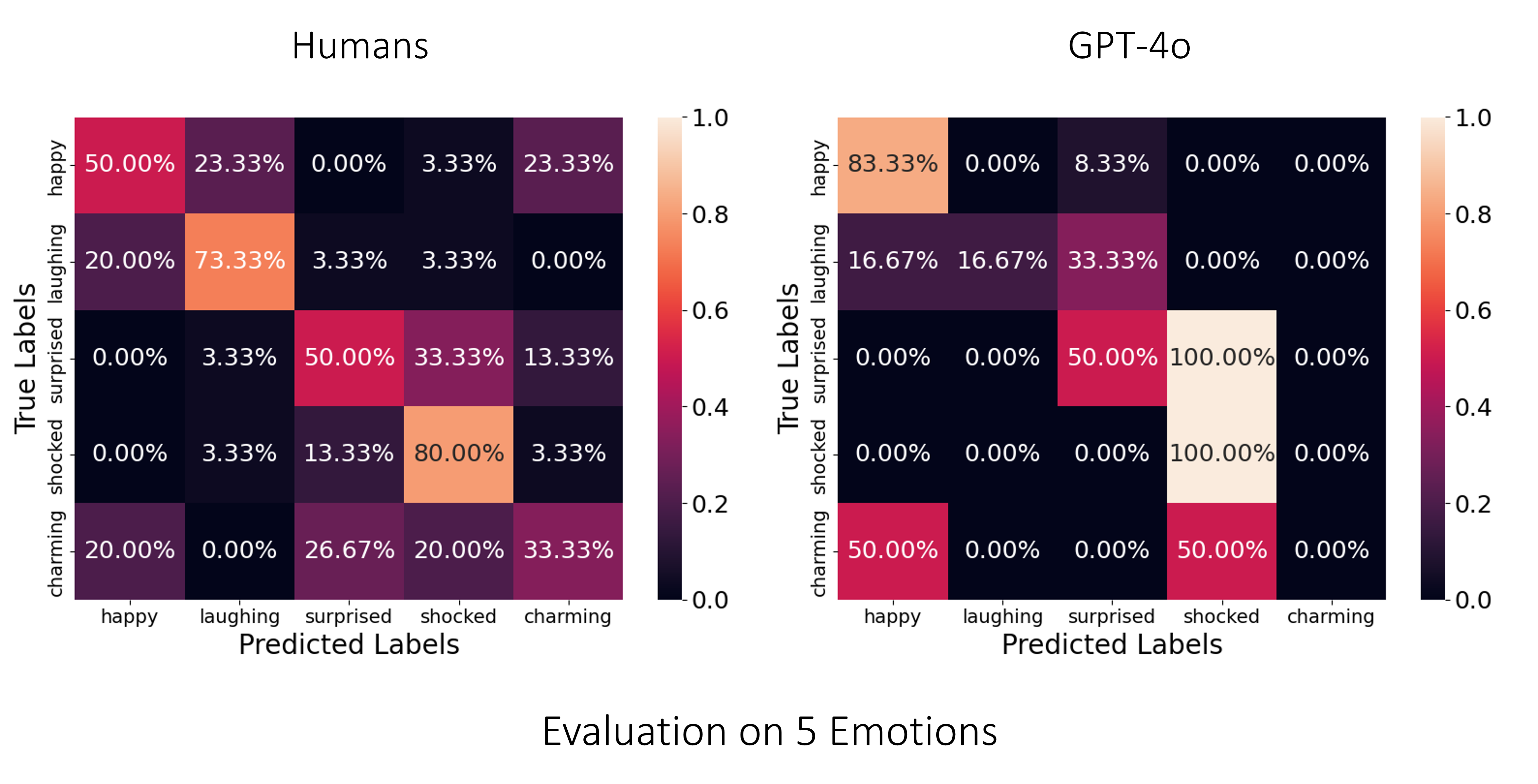}
    \end{minipage}
  \caption{Confusion matrices for (\textbf{Left}) evaluation on 3 emotions and (\textbf{Right}) evaluation on 5 emotions. The automatic evaluation shares a significant number of mistakes with the human annotators, indicating a reasonable level of emotional understanding even for our zero-shot scenario. Nevertheless, human evaluation can achieve significantly higher precision across the whole range of emotions.}
\label{fig:confusion_matrices_2D_3_and_5emotions}
\vspace{-1em}
\end{figure}

\vspace{2mm}
\noindent This methodology combines state-of-the-art techniques in pose estimation, depth perception, and controlled diffusion to create a novel approach to translating human emotions into artistic representations. The use of multiple control mechanisms (pose and depth) allows for nuanced control over the generated output, while the custom fine-tuning ensures consistency with a specific artistic style. The experimental setup involved applying this approach to the testing videos from our NED dataset and evaluating the quality of the transmitted emotion through the artistic generation in two scenarios with both manual and automatic evaluators. The evaluation was done with videos depicting only one of 3 emotions (happy, surprised, laughing) and a more challenging set of 5 emotions (happy, surprised, laughing, shocked, charming). We evaluate with both human annotators and automatic, multi-modal, LLM-based classification. 

\vspace{2mm}
\noindent The quantitative evaluation using both procedures is shown in Table \ref{tab:human-eval-diffusion}. Human annotators demonstrated remarkable proficiency, achieving a mean accuracy of 91.66\% in the three emotions scenario. This performance significantly surpassed that of GPT-4o, which managed an accuracy of 55.56\%. The substantial drop (36.1\%) underscores the human capacity to discern and categorize basic emotions with high precision. The confusion matrices (Figure~\ref{fig:confusion_matrices_2D_3_and_5emotions} (Left)) further illuminate this disparity, showing that humans achieved perfect accuracy in identifying "laughing" and "surprised" emotions, with only slight confusion between "happy" and "surprised" states. In contrast, GPT-4o struggled to differentiate between "happy" and "laughing," often conflating these two positive emotional states. Examining the confusion matrices for the scenario with five emotions (Figure~\ref{fig:confusion_matrices_2D_3_and_5emotions} (Right)) we noticed that human evaluators maintained relatively high accuracy for "laughing" and "shocked" emotions but displayed increased confusion across the board, particularly between similar emotional states like "happy" and "charming." GPT-4o, interestingly, showed perfect accuracy in identifying "shocked" emotions and high accuracy for "happy" states. However, it struggled with more subtle distinctions, completely failing to identify "charming" as a distinct emotion. While GPT-4o's overall performance was lower, its more consistent accuracy across both scenarios hints at a potentially scalable approach to emotion recognition. The model's ability to maintain relatively steady performance in the face of increased complexity is noteworthy, suggesting that with further refinement, AI models might offer valuable complementary tools in emotion recognition tasks, particularly in scenarios involving subtle or ambiguous emotional expressions.   

\subsection*{3D facial expression generation}
\vspace{2mm}
Apart from the 2D image generation, we also investigated ways to generate 3D content, first by exploiting facial expressions. We considered three methods for our 3D facial response pipeline. The first one uses PCA as a dimensionality reduction scheme, to create an embedded space for the pool of selected emotions. The second one is a neural network (NN) based on an unsupervised learning scheme that uses the signal from the first method to distill expressions, whilst the third is a classical, baseline procedure, based on retrieval from a pre-computed database of expressions. We apply our methods in the video spectrum, as finer-grained expressions are better caught considering both the space and time components. For the PCA method, we input 60 frames, for the NN we use 30 frames and for the baseline, retrieval method we use sequences of 60 frames. 

\subsubsection*{Expression Space Reconstruction}

Given an input video, we extract, at a frame-rate basis, the set of 146 three-dimensional facial keypoints, denoted by $(x, y, z)$ coordinates, using Mediapipe~\cite{lugaresi2019mediapipe}. For this method, in particular, we used 2 seconds as input, equivalent to 60 frames, and predicted the same amount of frames as the final result. 

We model spatiotemporal expressions or emotions using a sequence of keypoints in time and space as $N_{kp} \times T$, where $N_{kp}$ is the set of keypoints per frame and $T$ is the number of frames (moments in time). We define the \textit{expression vector} by $X_{i}$ with a dimension of $N_{kp} \times T \times 3$, for a given index $i$ in time. Our goal is that for a given expression $X_{i}$ we respond with a dedicated expression $X'_{i}$. For the simple case of mirroring, $X_{i} = X'_{i}$. As our baseline, we opted to compute PCA on a subset of samples from our dataset, for a predefined set of emotions, with the mean $X_{0}$ and $v_{k}$ as the set of principal components, for which we choose $k$ based on the magnitude of the eigenvalues $\lambda_{k}$. For choosing the best $k$, for a given set of emotions, we first sort the eigenvalues in decreasing order based on their magnitudes and choose $k^{*}$ such as:
\begin{equation}
    k^{*} = \arg\min_{k}\dfrac{\sum\limits_{i=1}^{k}\lambda_{i}}{\sum\limits_{i=1}^{N}\lambda_{i}} \geq 0.95
\end{equation}

For a given $X_{i}$ we compute the projection function on this space of emotions \(P(X_{i}) = X'_{i}\), where $X'_{i}$ will be the linear combination between all the selected eigenvectors \({v_{1}, v_{2}, ... v_{k*}}\) (the principal components).

\begin{equation}
    X'_{i} = X_{0} + \sum\limits_{j=1}^{k^{*}}(v_{j}^\intercal(X_{i} - X_{0}))v_{j}
\end{equation}

We denote by $c_{k^{*}} = v_{k^{*}}^\intercal(X_{i} - X_{0})$ the set of coefficients to which we add a small normal noise signal (to capture the improvisation element) to the embedded space of emotions represented by the PCA coefficients such that:
\begin{equation}
\label{eq:eig_w_noise}
    c'_{k^{*}} = c'_{k^{*}} + \sqrt{\lambda_{k^{*}}}\mathcal{N}(0, \sigma)
\end{equation}

In summary, the method does the following: 1) Learns the space of emotions in space (using facial keypoints) and time (from multiple video frames) using PCA; 2) Captures data and projects it on this space of emotions; 3) Produces a response by adding noise only in this embedded space (noise that is scaled by the $\sqrt\lambda$) for an element of improvisation and surprise.

\subsubsection*{Reaction Distillation}
\begin{figure*}[ht!]
\centering
\includegraphics[width=0.8\linewidth]{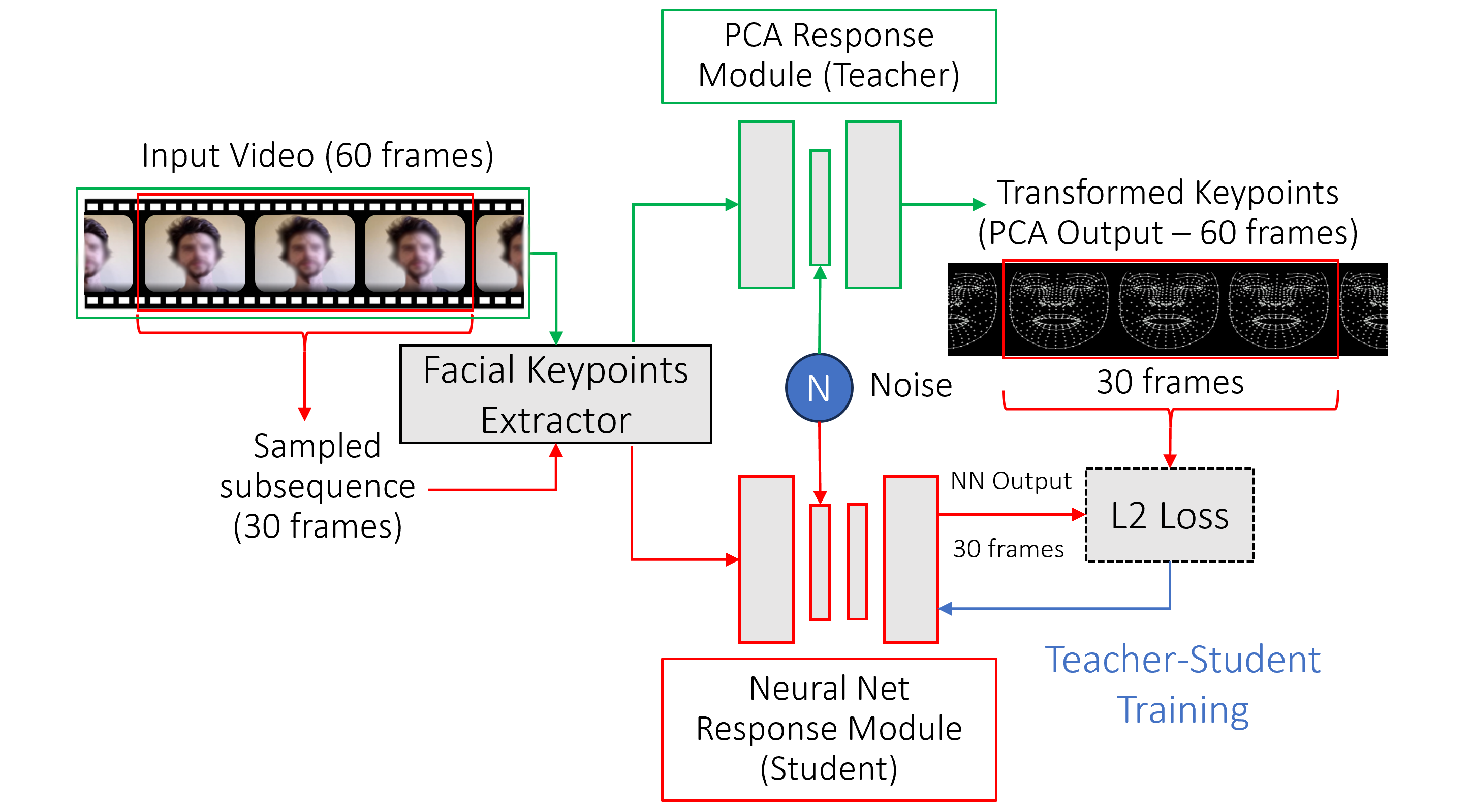}
\caption{Our unsupervised Teacher-Student training paradigm uses PCA as a supervisor to train a small NN for emotion distillation. Noise is added to the embedding space to convey an element of improvisation.}
\label{fig:main_figure}
\vspace{-1em}
\end{figure*}

\vspace{2mm}
One of our main contributions consists of building the first unsupervised learning framework for real-time reactive response based on the user's emotion, using a Teacher-Student paradigm. We propose a simple, MLP-based neural network for learning emotions, guided by the results from the embedded emotion space constructed with our PCA method. We designed this architecture to incorporate the same element of surprise found in the PCA algorithm - the $ \sqrt{\lambda_{k^{*}}}\mathcal{N}(0, \sigma)$ vector in Equation~\ref{eq:eig_w_noise}. Different from the PCA, the keypoint sequence length is shorter - 30 frames compared to the 60 frames for the PCA. We argue that by being more powerful, the neural net could learn from fewer samples and would provide a faster response for a more natural nonverbal interaction. The final architecture is made from fully connected layers, the first that brings the dimension down to match the noise vector from PCA, a second one with the same number of features as the noise size, and a final one that generates the same sequence length as the input - 30 frames. The full training pipeline is presented in Figure~\ref{fig:main_figure}. 

\subsubsection*{Similar Emotion Retrieval}
\vspace{2mm}
Given an input video, we retrieve the most similar keypoint sequence (2 seconds in length) from a predefined dataset of videos of different emotions conveyed using facial expressions (saved in the form of facial keypoints) as the reaction. The training dataset is sampled every 30 frames and for each 30 frames of testing video, a new sequence is presented as the response from the searched training space. The setup is similar to PCA - 146 input keypoints, 60 frames sequence for both input and output. For each sequence, the mean is subtracted from both the sample and training set. Retrieval is the closest method to mirroring that we propose and also one of the fastest. Nevertheless, it lacks the novelty brought by noise from the dimensionality reduction and learning schemes. 

\vspace{2mm}
\noindent These methods were evaluated (Table~\ref{tab:human-eval}) by both human annotators and the GPT-4o model, focusing on their ability to generate and convey three basic emotions: happy, laughing, and surprised. 
The PCA method emerges as the most successful approach when evaluated by human annotators, which makes it a suitable candidate as a supervisor for the learning approach. The Teacher-Student paradigm was also effective and valid since the NN method comes second. The fact that the NN-Student did not surpass the PCA-Teacher can be explained by the way we opted to train the model --- in order to increase the number of training samples we used 30 frames input rather than 60 frames from the PCA sequences. We argue that more training data could also translate to better results and eventually the Student exceeding the Teacher. The simplicity of the Retrieval procedure, which is a Nearest Neighbor method in the spatiotemporal space of sequences of keypoints, translates to poor results. 
Interestingly, the GPT-4o model's evaluation presents a different ranking of the methods. The Retrieval approach scores highest at 38.89\%, followed by the NN at 27.78\%, and PCA at 22.22\%. This reversal in performance ranking between human and AI evaluators highlights the challenges in developing AI systems that can accurately perceive and interpret human emotions in the same way humans do. The discrepancy between human and GPT-4o evaluations is particularly pronounced for the PCA method, which performs best with humans but worst with GPT-4o. This divergence may indicate that the PCA method produces expressions that are more naturalistic or subtle in ways that are easily perceived by humans but not readily captured by the current AI model.



\begin{table}
    \setlength{\tabcolsep}{2.5pt} 
    \centering
    \begin{tabular}{lcc}
        \toprule
        Method & Humans (Mean) & GPT-4o\\
        \toprule
        NN & 55.56 & 27.78 \\
        PCA & \textbf{66.67} & 22.22 \\
        Retrival & 44.44 & 38.89\\
        All & 55.56 & 29.63\\
        \bottomrule
    \end{tabular} 
    \caption{Human and machine-level evaluation for each of our proposed methods and overall. We present the accuracy over all classes in percentages for both human-level evaluation and automatic evaluation. The best result is bolded.}
    \label{tab:human-eval}
\vspace{-1em}
\end{table}


\begin{figure*}[ht!]
\centering
\includegraphics[width=0.8\linewidth]{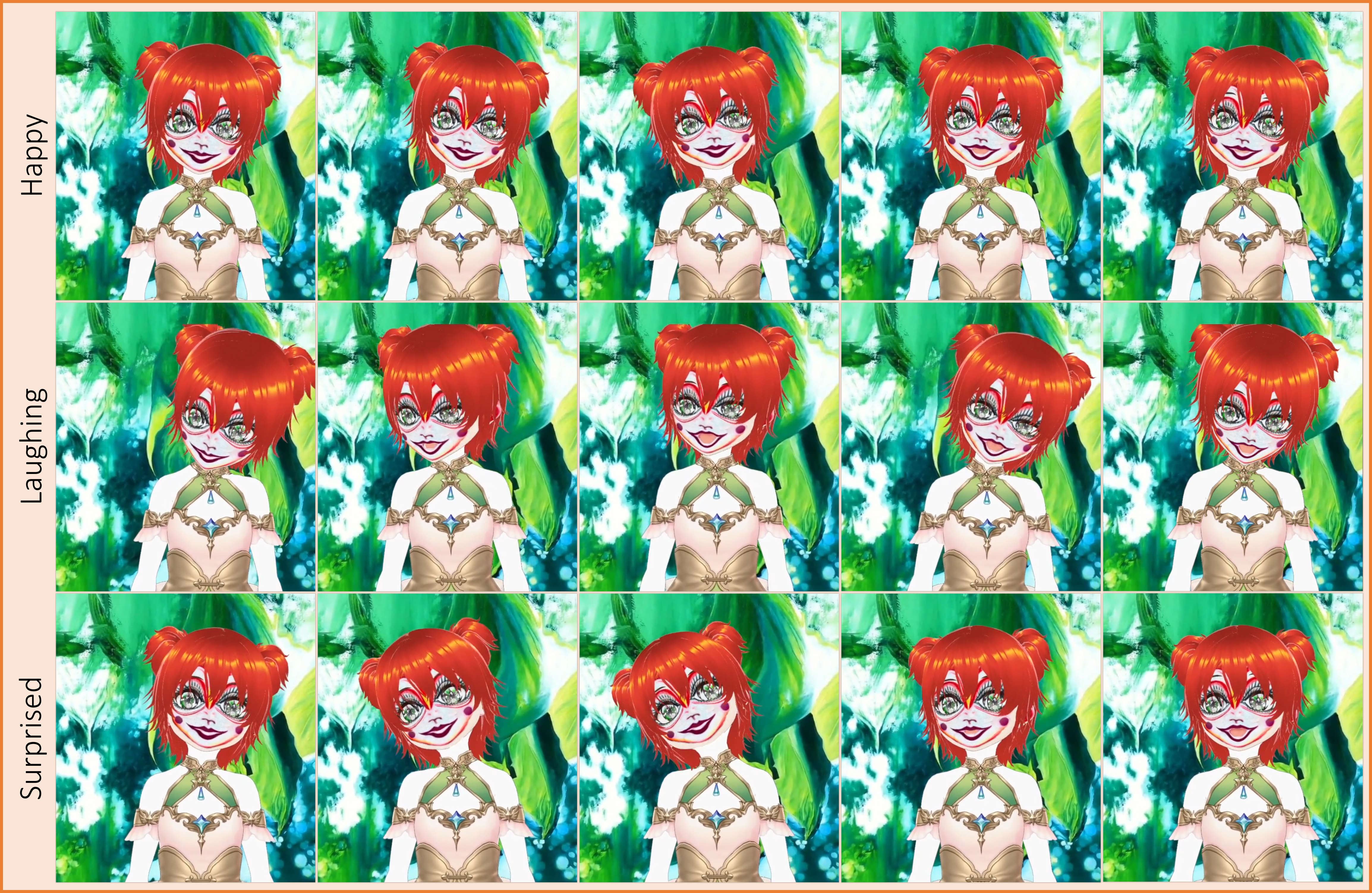}
\caption{Maia in action. We present visual results of our PCA method applied to our 3D avatar. Currently, we focused on 3 positive emotions but can be extended to many more. We can clearly see a natural and expressive response to all emotions and a friendly demeanor, perfectly suited for human interaction, especially with children.}
\label{fig:qual_results_pca}
\vspace{-1em}
\end{figure*}

\vspace{2mm}
\noindent In Figure~\ref{fig:qual_results_pca} we present qualitative results of our friendly 3D avatar Maia expressing emotions of happiness, laughter and surprise, responding to the user's emotion, after applying our PCA method, which acts as the Teacher for the Student NN method. We can clearly see how naturally and human-like the expressions of our character are through multiple nonverbal cues --- for \textit{happy}, Maia often displays a soft smile, with the corners of the mouth turning upward and lifting cheeks, along with bright eyes and relaxed or cheerful posture --- \textit{laughing} is characterized by a wider opened mouth, and is often accompanied by a rhythmic and dynamic shaking of the head and body, while \textit{surprised} is typically conveyed through widened eyes, raised eyebrows, and a slightly open mouth, signifying astonishment or disbelief. 
Each interaction with Maia is personalized, derived from the user's mode of interaction and is also unique, the response is never the same, because of the element of surprise we injected in the pipeline, which makes it more engaging and fun.


\begin{figure*}[ht!]
\centering
\includegraphics[width=0.8\linewidth]{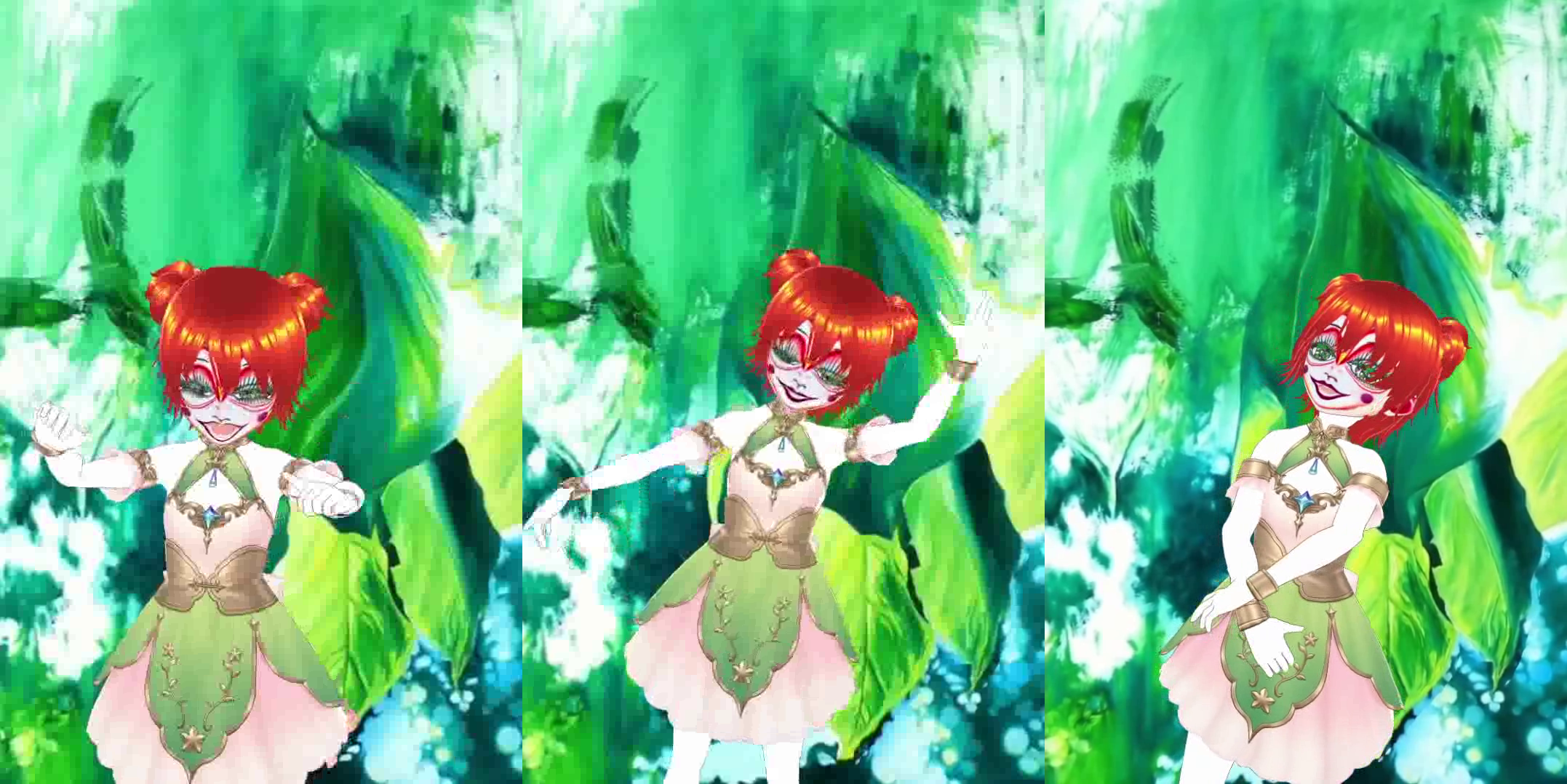}
\caption{Maia - body in action. We compiled a set of common interactions with the avatar. This resulted in three emotions, \textit{happy to see you}, \textit{enthusiastic} and \textit{laughing}. Body movements further enhance the expressiveness of the character.} 
\label{fig:qual_results_body}
\vspace{-1em}
\end{figure*}


\subsection*{3D body movement generation}
\vspace{2mm}
\noindent We recently showcased Maia at NEOArt Connect~\cite{Maia_NeoArtConnect}, the largest art and science exhibition in Romania and at the prestigious art gallery Galateca in Bucharest \cite{Maia_Galateca}. These art and science events, having a large and diverse public attendance, provided the opportunity to gather real-world feedback from hundreds of participants, of different age, gender and background, who engaged in interacting with Maia. Before each interaction, we obtained informed consent from users, clearly explaining Maia's functionality, data collection processes, and its one-on-one interaction capability. 

\vspace{2mm}
\noindent We have projected Maia on a large screen and deployed her using our real-time interaction pipeline for the first nonverbal chat. We captured a continuous data stream (video) with people's interaction with Maia. This input video stream is sampled into 30 or 60-frame chunks that are fed to Mediapipe~\cite{lugaresi2019mediapipe} for face detection. When multiple individuals are in the range of the camera capturing the interaction, we focus on the interaction of the person closest to the center of the frame. Afterwards, we extracted the set of 146 facial keypoints and then transformed them using one of our methods (PCA-Teacher, NN-Student, Retrieval-Baseline) and used to animate Maia using VSeeFace. Given the lightweight nature of the transforms, the pipeline runs in real-time even on a portable device. In this iteration, we also incorporated body movements. Maia's body movements were not directly mirrored but rather responded to human movements in a manner consistent with Maia's expressed emotions. The response from the individuals interacting with Maia was overwhelmingly positive, with hundreds of participants engaging with Maia. Interaction durations ranged from 4 to 27 minutes per participant, demonstrating the system's capacity to maintain prolonged, enjoyable engagement. This enthusiastic reception serves as a testament to the compelling nature of our work and its ability to captivate users. 

\vspace{2mm}
\noindent Motivated by this positive response, we conducted a new set of experiments involving body movements, through more keypoints (not just facial), without employing any learning procedure as before, just by mimicking the user's interaction with Maia. We focused on three emotions: "happy to see you", "enthusiastic," and "laughing," as these were the most frequent modes of interaction with Maia (samples from the dataset can be seen in Figure~\ref{fig:qual_results_body}). For each of these emotions we collected around 100 videos of 5 seconds each, on average, and we gave our evaluators (human and GPT-4o) these videos to identify how well Maia as a character conveyed the desired emotion through body movement. 

\begin{figure}[ht!]
  \centering
  \includegraphics[width=1.02\linewidth]{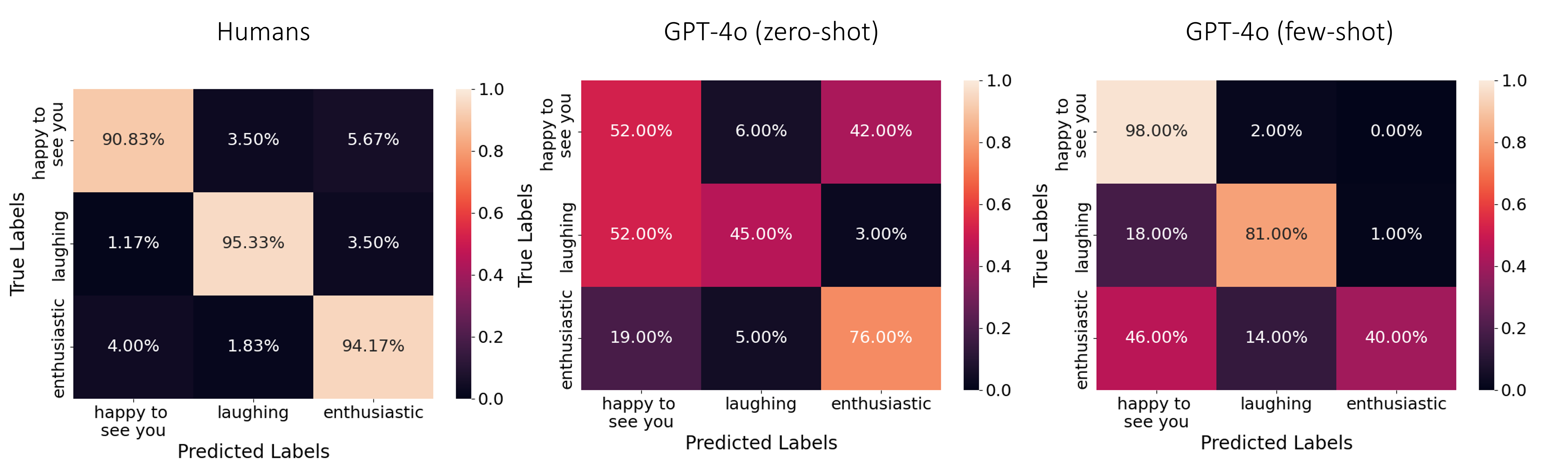}
  \caption{Confusion matrices for Maia-body dataset. From left to right, human evaluation, GPT-4o zero-shot and GPT-4o few-shot. Even with several training samples seen (few-shot), GPT-4o yields poorer performance compared to the human evaluation.}
\label{fig:confusion_matrices_body}
\end{figure}

\vspace{2mm}
\noindent Compared to our previous zero-shot automatic evaluation procedures, for these experiments exclusively, we employ a few-shot evaluation procedure for GPT-4o where we give samples from each class before giving the test videos for evaluation. For fair comparison, humans also received the same samples per class, before starting the evaluation procedure. Quantitative results are reported in Table~\ref{tab:human-eval-body} and Figure~\ref{fig:confusion_matrices_body}. Human evaluators demonstrated remarkable proficiency in identifying the intended emotions, achieving a mean accuracy of 93.33\%. This high accuracy underscores the effectiveness of the body movement generation method in conveying emotions that are easily interpretable by humans. In contrast, the GPT-4o model's performance varied significantly between the zero-shot and few-shot approaches. In the zero-shot scenario, where the model had no prior examples to learn from, GPT-4o achieved an accuracy of 57.67\%. While this is considerably lower than human performance, it still indicates some ability to discern emotional cues from body movements without specific training. The confusion matrix for the zero-shot evaluation reveals that GPT-4o struggled most with distinguishing between "happy to see you" and "enthusiastic," often confusing these two positive emotions. The few-shot approach, where GPT-4o was provided with sample videos for each emotion class before evaluation, showed a marked performance improvement, reaching an accuracy of 73.00\%. The confusion matrix for the few-shot evaluation shows significant improvements across all emotions, particularly for "happy to see you" (98.00\% accuracy) and "laughing" (81.00\% accuracy). However, the model still showed some difficulty in accurately identifying "enthusiastic" emotions, often misclassifying them as "happy to see you". The progression from zero-shot to few-shot evaluation in GPT-4o's performance is particularly noteworthy. It suggests that with even a small amount of targeted training or context, AI models can significantly enhance their ability to interpret complex, nonverbal emotional cues. These results validate the effectiveness of the Maia body movement generation method in creating distinct and recognizable emotional expressions. The high accuracy achieved by human evaluators suggests that the generated movements by Maia successfully capture key characteristics of each emotion.


\begin{table}
    \setlength{\tabcolsep}{2.5pt} 
    \centering
    \begin{tabular}{lccc}
        \toprule
        Set & Humans (Mean) & GPT-4o zero-shot & GPT-4o few-shot\\
        \toprule
        3 emotions & \textbf{93.33} & 57.67 & 73.00 \\
        \bottomrule
    \end{tabular} 
    \caption{Human-level and automatic evaluation for the Maia body movement experiments on our test set.}
    \label{tab:human-eval-body}
\end{table}



\subsection*{Artistic considerations} 
\vspace{2mm}

Our nonverbal AI chat system represents a groundbreaking fusion of technology and art, embodying Ray Kurzweil's concept of art as a form of human technology~\cite{kurzweil2005singularity}. This innovative approach establishes a novel Human-AI interaction paradigm rooted in emotional exchange, where the machine's capacity to "feel" is manifested through its ability to transmit and respond to emotional cues. The system's uniqueness lies in the expressive interplay between human facial expressions, body poses and AI-generated artistic output, transcending mere aesthetics or AI-generated art combinations to create a new art form born from real-time emotional Human-AI interaction. This dynamic intersection of human artistry, facial expressions, gestures, and algorithmic creation aligns with Rudolf Arnheim's observations on the clarity and precision required in visual communication \cite{arnheim1954art}. In Maia, we strive for this clarity, ensuring that the AI's responses are as expressive and precise as possible, transforming computer vision outputs into meaningful artistic expressions. The digital brushstrokes of our AI-generated responses (Figure~\ref{fig:maia_evolution} second row) mirror the nuanced articulation of human emotions, fostering a unique dialogue between humans and machines. This innovative approach not only pushes the boundaries of Human-AI interaction but also contributes to the evolving landscape of digital art and emotional communication, challenging our perceptions of the limits between human and machine creativity and inviting exploration of new expressive forms in the digital age. 
Maia opens new avenues for understanding and exploring emotional expression and perception, potentially enriching our comprehension of human emotional communication and paving the way for future advancements in the field of AI-assisted artistic creation and emotional interaction.

\subsection*{Ethical considerations}
\vspace{2mm}
Privacy and data protection are paramount in our approach to developing Maia. We have implemented robust measures to safeguard user information while maintaining the effectiveness of our solution. Our method's reliance on facial and body keypoints, rather than raw video data, serves as a critical privacy-preserving feature. During the training process, we extract these keypoints from facial data and emotional insights, anonymizing the information and discarding the original video input. This approach allows us to build a comprehensive data lake of emotions from multiple interactions over time without risking user privacy. Furthermore, our focus on motion patterns correlated with specific emotions, rather than demographic factors such as gender, race, or age, minimizes potential biases and enhances the system's adaptability. As we continue to refine Maia, we recognize the need to develop domain-specific ethical guidelines, particularly for sensitive applications in fields like therapy and education. Nevertheless, a demographically diverse dataset is also one of our objectives that will be reflected in future iterations of this work.


\subsection*{Impact}   
\vspace{2mm}

Nonverbal interaction is the first form of interaction infants learn (nonverbal engagement between a mother and her child) - first by mimicking and then alternating their response based on the mother’s expression --- learning of emotions. Special needs children require a special form of interaction and permanent engagement --- qualified personnel do not meet current demands, as reported by the World Health Organization~\cite{krahn2011world}. Exploring the societal impact of our nonverbal communication system, we envision its application not just in entertainment but as a transformative tool in education, therapy for individuals with communication challenges, and interactive museum exhibits. By facilitating intuitive and engaging Human-AI interactions, our system holds the potential to revolutionize how we educate, entertain, and assist diverse communities, especially benefiting those with communication challenges.


\section*{Conclusions}
\vspace{2mm}

Our pioneering real-time nonverbal chat system uses facial keypoints to interpret and respond to a wide range of emotions, with potential for incorporating body movements for more comprehensive interaction. It aims to enhance AI communication by addressing the limitations of current Large Language Models (LLMs), which excel in text-based tasks but cannot engage in complex, multi-modal interactions that characterize human communication. While LLMs are trained on vast datasets, they struggle to tailor responses to individual needs and miss crucial non-verbal cues like gestures, body language, and facial expressions. These elements convey complex emotional and intentional information beyond the written text. By focusing on these nonverbal aspects, the system seeks to bridge the gap between AI capabilities and the nuanced nature of human interaction, potentially leading to more intuitive and comprehensive Human-AI communication that goes beyond the constraints of text-based exchanges.

\vspace{2mm}
\noindent By bridging the gap between verbal and nonverbal AI communication, our work represents a significant leap forward in the field of Human-AI interaction. We believe that Maia opens new avenues for creating more empathetic, engaging, and naturalistic AI systems, bringing us one step closer to the goal of truly intuitive human-machine interfaces through Art.

\section*{Additional information}
\noindent\textbf{Competing interests}
All authors declare that they have no conflicts of interest.

\noindent\textbf{Data collection and experimental protocols}
All data used in the experiments consists of RGB videos, where each video presents a person conveying an emotion, performing certain body movements, gestures and facial expressions. Each participant gave their consent for the collection and storage of the data for the scope of the dataset and research, in accordance to GDPR regulations. 
All our experimental protocols were approved by the official Ethics Committee of the University Politehnica of Bucharest, which is the institution where all our research and experiments took place.

\noindent\textbf{Data availability}
All the data, which we make available, includes the raw head and body videos, processed as described in the paper. All videos used for the experiments in this paper are available at the following link:

\url{https://sites.google.com/view/spacetime-vision-robotics-lab/maia}

\noindent\textbf{Author contribution}
All authors (DC, AM, CL, ML) contributed to the conceptualization and manuscript writing. Technical methodology was developed by DC, AM, and ML. Artistic works, including both physical and digital character creation, were contributed to by all authors. Data collection was performed collaboratively by DC, CL, and ML. System implementation and development was carried out from all authors. 
All authors have read and agreed to the present version of the manuscript.

\subsection*{Acknowledgements} 
This work was supported in part by the EU Horizon project ELIAS (Grant number: 101120237) and by UEFISCDI through EEA and Norway Grant 2019-2022: EEA-RO-2018-0496. We want to express our deepest gratitude to Andreea Sandu and Anca Boeriu, the curators and organizers of NeoArt Connect and Galateca Art Gallery, for inviting us to exhibit Maia at their prestigious events. We also want to express our most sincere gratitude towards Aurelian Marcu from NILPRP for providing access to GPU resources from the National Interest Infrastructure facility IOSIN-CETAL.
{\small
\bibliography{main}
}

\end{document}